\documentclass[11pt]{article}
\PassOptionsToPackage{table,dvipsnames}{xcolor}
\usepackage[final]{acl}

\usepackage{times}
\usepackage{latexsym}

\usepackage[T1]{fontenc}

\usepackage[utf8]{inputenc}

\usepackage{microtype}

\usepackage{inconsolata}

\usepackage{graphicx}

\usepackage{xcolor}         
\usepackage{wrapfig}
\usepackage{tabularx} 
\usepackage{booktabs} 
\usepackage{array}    
\usepackage{array}
\usepackage{colortbl}
\usepackage{makecell}

\usepackage{enumitem}

\usepackage{graphicx}

\usepackage{amsmath}  
\usepackage{amsthm}   
\usepackage{algorithm}
\usepackage{algpseudocode}

\usepackage{mathtools}
\usepackage{amsthm}
\theoremstyle{plain}
\newtheorem{finding}{Finding}

\usepackage{longtable}      
\usepackage{booktabs}       
\usepackage{multirow}       
\usepackage{array}          
\usepackage{tabularray}      
\UseTblrLibrary{booktabs}   

\definecolor{headerblue}{HTML}{E8F4FD} 
\definecolor{rowgray}{HTML}{F5F5F5}    

\definecolor{deltagreen}{RGB}{0,140,0}

\newcommand{\scoredelta}[2]{%
  \begingroup
  \setlength{\tabcolsep}{0pt}%
  \begin{tabular}{@{}l@{}l@{}}%
    #1 & {\tiny\color{deltagreen}\hspace{0.15em}$#2$}%
  \end{tabular}%
  \endgroup
}

\definecolor{headerblue}{HTML}{E8F4FD} 
\definecolor{rowgray}{HTML}{F5F5F5}    

\usepackage[most]{tcolorbox}
\usepackage{float}
\usepackage{xspace}
\tcbset{
aibox/.style={
width=450pt,
top=8pt,
colback=white,
colframe=black,
colbacktitle=black,
enhanced,
center,
attach boxed title to top left={yshift=-0.12in,xshift=0.15in},
boxed title style={boxrule=0pt,colframe=white},
}
}
\newtcolorbox{AIbox}[2][]{aibox,title=#2,#1}\tiny

\title{ClinAlign: Scaling Healthcare Alignment from Clinician Preference}

\author{
\textbf{Shiwei Lyu\textsuperscript{1}\thanks{ Equal Contribution.}}
\textbf{, Xidong Wang\textsuperscript{1,2}\footnotemark[1]},
\textbf{Lei Liu\textsuperscript{1}},
\textbf{Hao Zhu\textsuperscript{1}},
\textbf{Chaohe Zhang\textsuperscript{3}},
\\
\textbf{Jian Wang\textsuperscript{1}},
\textbf{Jinjie Gu\textsuperscript{1}},
\textbf{Benyou Wang\textsuperscript{2}},
\textbf{Yue Shen\textsuperscript{1}}
\\
\textsuperscript{1}Ant Group
\textsuperscript{2}The Chinese University of Hong Kong, Shenzhen
\textsuperscript{3}Peking University
\\
\small{
    \textbf{Email: }{ lvshiwei.lsw, zhanying@antgroup.com}
}\\
\url{https://github.com/AQ-MedAI/ClinAlign}
}

\begin{document}
\maketitle
\begin{abstract}

Although large language models (LLMs) demonstrate expert-level medical knowledge, aligning their open-ended outputs with fine-grained clinician preferences remains challenging. Existing methods often rely on coarse objectives or unreliable automated judges that are weakly grounded in professional guidelines. We propose a two-stage framework to address this gap. First, we introduce \textbf{HealthRubrics}, a dataset of 7{,}034 physician-verified preference examples in which clinicians refine LLM-drafted rubrics to meet rigorous medical standards. Second, we distill these rubrics into \textbf{HealthPrinciples}: 119 broadly reusable, clinically grounded principles organized by clinical dimensions, enabling scalable supervision beyond manual annotation. We use HealthPrinciples for (1) offline alignment by synthesizing rubrics for unlabeled queries and (2) an inference-time tool for guided self-revision. A 30B-A3B model trained with our framework achieves 33.4\% on HealthBench-Hard, outperforming much larger models including Deepseek-R1 and o3, establishing a resource-efficient baseline for clinical alignment.

\end{abstract}

\section{Introduction}

Healthcare is a high stakes domain where language models are increasingly deployed as clinical assistants. Scaling model capacity and medical corpora has driven large gains on knowledge intensive, exam style benchmarks \citep{yang2022gatortron,dou2025baichuan}. As these gains plateau, the key challenge shifts to aligning open ended responses with clinician preferences and professional standards in real consultations~\citep{fast2024autonomous, qiu2025evolving}. General RLHF goals such as helpfulness, honesty, and harmlessness are too coarse for clinical use\citep{johri2025evaluation,zhou2025large}, where desired behavior depends on urgency, uncertainty, and user expertise, motivating fine grained, instance specific rubrics.


\begin{figure}[t]
    \centering
    \includegraphics[width=1.0\columnwidth]{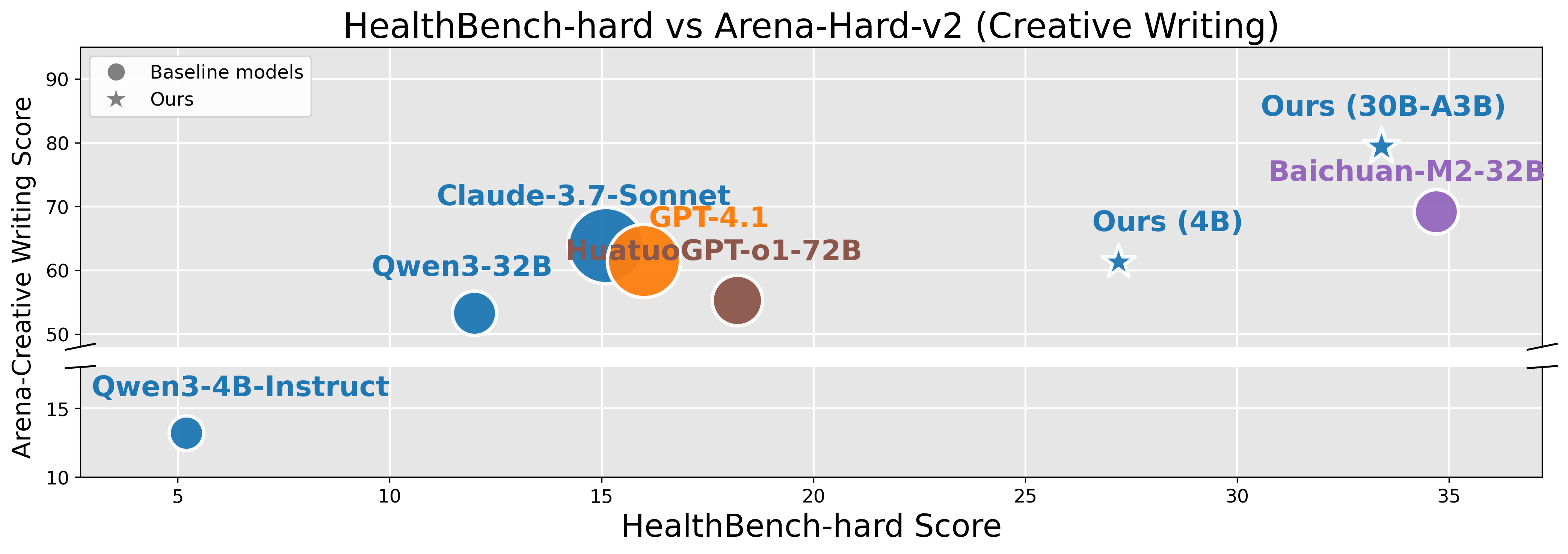}
    \vspace{-5mm}
    \caption{Scatter plot of performance where the x-axis shows the HealthBench-hard score and the y-axis shows the Arena-Hard-v2 Creative Writing score. Marker size is proportional to the model parameter count.}
    \vspace{-5mm}
    \label{fig:scaling}
\end{figure}

In response, evaluation is moving from \textit{exam-style tests}~\citep{zuo2025medxpertqa} to \textit{scenario-grounded assessment}. For example, HealthBench~\citep{arora2025healthbench} foregrounds rubric based scoring over realistic prompts, and rubric driven optimization increasingly uses such criteria for training~\citep{fast2024autonomous, team2025kimi}. However, these efforts have not yet yielded a  scalable approach to learning from clinician rubric expertise, largely due to scarcity of  clinician rubric data and expensive cost of clinician participation. Early efforts such as InfiMed~\citep{wang2025infimed} derive evaluation rubrics automatically from benchmark seeds, which can lead to overfitting to the benchmarks and limited generalization across clinical scenarios.
RaR-Medicine~\citep{gunjal2025rubrics} primarily focuses on exam-style question answering, without grounding evaluation in real-world physician expertise.




In this work, we curate a physician supervised preference dataset from real medical queries and multi model responses \citep{chiang2024chatbotarenaopenplatform,wang2025helpsteer3preferenceopenhumanannotatedpreference}. We drafted candidate rubrics using \texttt{GPT-5.1}, which were then selected, revised, and extended by physicians, resulting in 7,034 physician-verified supervision examples. Training \texttt{Qwen3-4B-Instruct} on this data raises HealthBench Hard from 5.2\% to 22.9\%, surpassing \texttt{GPT-5.1-Instant} at 20.8\%. However, per instance physician rubric authoring is costly and hard to scale to long tail scenarios.


To scale beyond per-instance annotation, we distill recurring rubric patterns into a reusable library of scenario-specific consensus rubrics, termed \textbf{principles}. We curate 119 principles with physicians, organized by urgency, uncertainty, user expertise, and task type. These principles enable rubric-grounded supervision for new real-world questions at scale, adding 16,872 examples, and are further packaged as an alignment tool that provides rubric references for inference-time self-revision. As shown in Figure~\ref{fig:scaling}, our models favorably against both open- and closed-source baselines on HealthBench-hard~\citep{arora2025healthbench} and Arena-Hard-v2~\citep{chiang2024chatbotarenaopenplatform}, demonstrating strong cross-benchmark performance without increasing model size.

{Our main \textbf{contributions }are threefold.}
\textit{(i)} We introduce \textbf{HealthRubrics}, a physician-verified dataset of 7,034 examples, constructed by having clinicians select, edit, and extend LLM-drafted instance-level rubrics, and demonstrate its effectiveness for alignment training; the dataset will be publicly released upon paper acceptance.
\textit{(ii)} We propose \textbf{HealthPrinciples}, a set of 119 reusable principles for scalable synthetic data generation, enabling models to outperform substantially larger commercial and open source systems on both HealthBench-Hard and Arena-Hard-v2.
\textit{(iii)} We develop a principle-driven self-revision mechanism for inference, achieving consistent improvements through online rubric-guided refinement.

\section{Related Work}
\paragraph{From Medical LLMs to Clinical Agents}
Progress in large language models has advanced healthcare along two complementary directions. On the modeling side, both proprietary and open medical LLMs have achieved strong performance on knowledge-intensive evaluations, covering standardized exams, unstructured clinical text~\citep{yang2022gatortron}, Chinese medical corpora~\citep{wang2025huatuo}, and lightweight task-oriented designs~\citep{wang2024apollo}. Reinforcement learning further improves medical adaptation and reasoning~\citep{dou2025baichuan,chen2024huatuogpt}. On the systems side, agentic clinical assistants extend beyond single-turn generation via multi-step reasoning, tool-augmented interaction~\citep{zhao2025agentic}, and integration with structured medical resources, enabling applications such as differential diagnosis~\citep{qiu2025evolving} and radiology reporting~\citep{oh2024llm}. Despite these advances, aligning open-ended responses with clinician preferences and professional standards across diverse clinical contexts remains underexplored. We address this gap with practical data and method baselines for fine-grained clinician preference alignment.

\begin{figure*}[t]
    \centering
    \hspace*{-0.16cm}%
    \includegraphics[width=2.05\columnwidth]{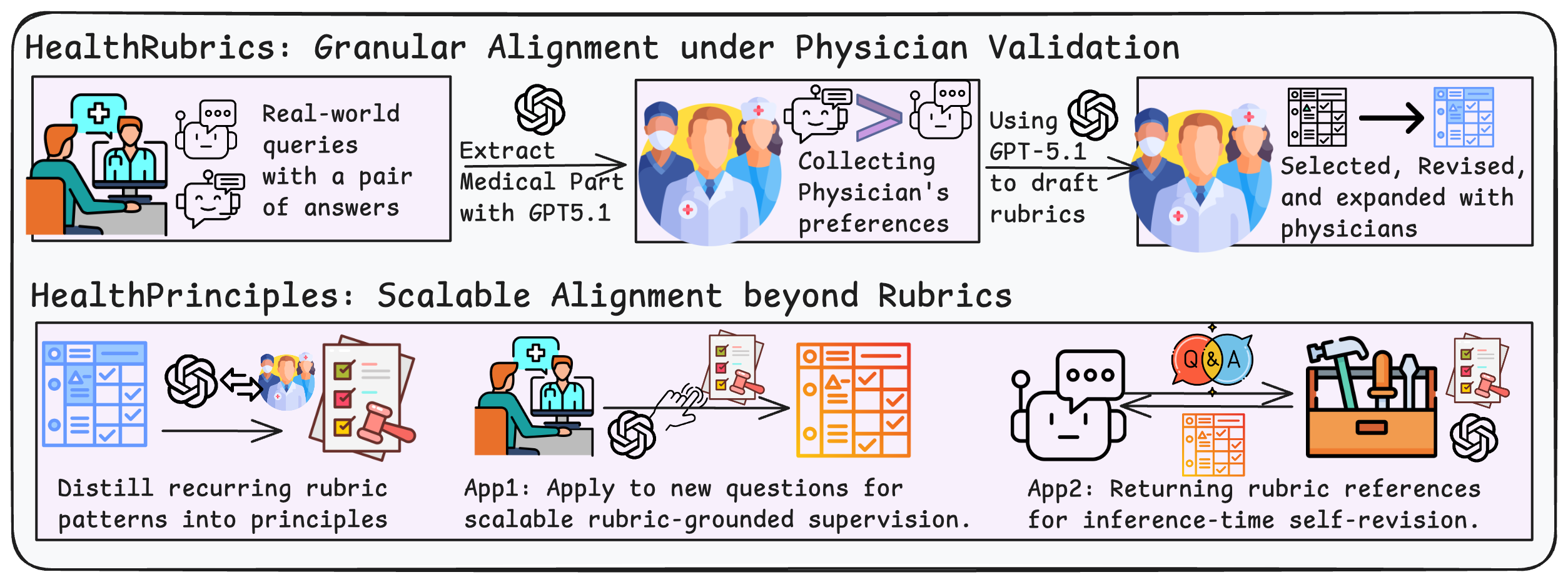}
    \vspace{-1mm}
    \caption{\textbf{Method overview.} \textbf{(Top) HealthRubrics:} we draft rubrics with \texttt{GPT-5.1} for real-world medical queries and multi-model responses, then have physicians refine them into validated preference supervision. \textbf{(Bottom) HealthPrinciples:} we distill recurring rubric patterns into reusable, scenario-specific principles, used to (i) scale rubric-grounded supervision to new questions and (ii) provide rubric references for inference-time self-revision.}
    \vspace{-3mm}
    \label{fig:method}
\end{figure*}

\paragraph{Towards Rubric Evaluations}
Medical LLM evaluation has moved beyond multiple-choice knowledge tests toward scenario-grounded assessment, as early benchmarks are easy to score but fail to capture critical consultation behaviors such as long-form reasoning, communication quality, and safety compliance~\citep{zuo2025medxpertqa,wang2024cmb}. Recent efforts emphasize real-world data and more reliable automated pipelines. LLMEval-Med~\citep{zhang2025llmeval} derives cases from EHRs and expert-designed scenarios, combining expert checklists with LLM-based judging refined through human agreement. HealthBench~\citep{arora2025healthbench} further scales to realistic health conversations for both lay users and clinicians, using example-specific rubrics and multi-axis scoring to evaluate long responses and diverse behaviors. However, these approaches remain largely evaluation-centric: rubrics and checklists are rarely reused as supervision for training, limiting scalable preference alignment.

\paragraph{Rubric RL}
Rubric based evaluation decomposes complex goals into verifiable criteria, providing structured and interpretable supervision across domains~\citep{scale2025vista,lin2024wildbench,starace2025paperbench,fast2024autonomous}. Recent work uses rubrics as reward or preference signals, often with LLM graders~\citep{team2025kimi}, and sometimes incorporates rubric guidance into rollout and policy learning~\citep{gunjal2025rubrics,zhou2025breaking,jayalath2025compute}. These approaches face recurring challenges, including objective conflicts and reward hacking from poorly constrained rubrics, and high cost and instability from online judging~\citep{eisenstein2023helping,fu2025reward}. In healthcare, follow up work extends rubric scoring from evaluation to training, including incremental RL for medical dialogue~\citep{wang2025infimed,gunjal2025rubrics,jin2025multidimensional}. Self refinement methods provide complementary self feedback signals, and interactive diagnostic agents use physician validated rubrics to evaluate diagnostic trajectories~\citep{qiu2025evolving,zhou2025enhancing}. However, existing pipelines either rely heavily on online judges or produce rubrics that are difficult to reuse across scenarios. Our work is distinguished by physician supervised rubrics and reusable consensus principles that support both scalable offline training and rubric referencing at test time.





\section{Pilot Study: Generalization Failure of Naive SFT}
\label{sec:pilot}



\textit{Healthcare alignment} refers to aligning a model's clinical responses with professional standards and clinician preferences in a context-aware, safety-critical manner that generalizes across real-world medical scenarios. In this section, we examine whether vanilla supervised fine-tuning (SFT) can achieve such alignment, using HealthBench as the evaluation benchmark.

\paragraph{Settings.}
We ask a basic question: \textit{Can SFT achieve fine-grained healthcare alignment?} We randomly split HealthBench into 3{,}000 training questions and 2{,}000 held-out questions. To construct SFT data, we use a strong contemporary LLM, \texttt{GPT-5.1}, to generate three rubric-aware responses per training question, conditioned on the question, the rubrics, a model draft, and the clinician ideal completion. This yields 9{,}000 training instances; the prompt template is provided in Appendix~\ref{app:sftprompt}. We then fine-tune \texttt{Qwen3-4B-Instruct} for two epochs using AdamW with a learning rate of $1\times10^{-5}$, a cosine schedule with 3\% warmup, and a global batch size of 64.

\begin{figure}[ht]
    \centering
    \includegraphics[width=1.0\columnwidth]{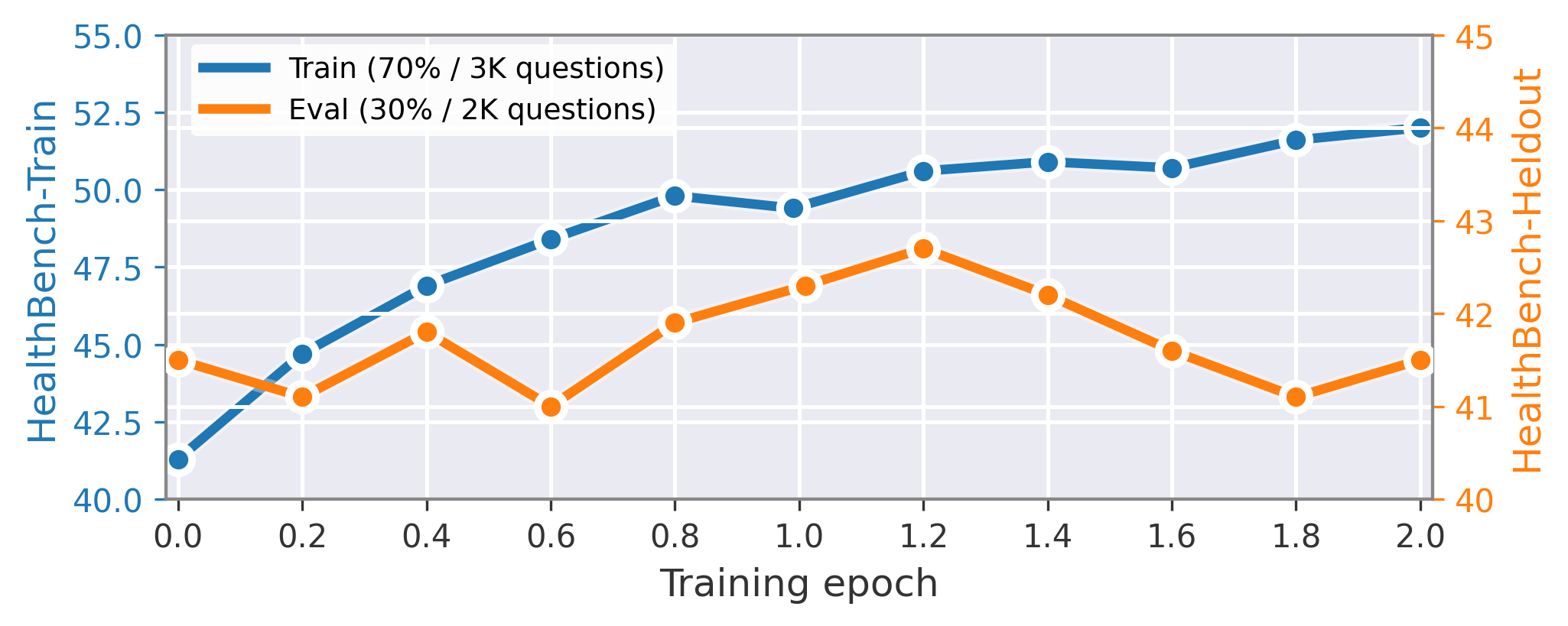}
    \caption{HealthBench scores vs. training epoch on a random 70/30 split with 3K training questions and 2K held-out questions, evaluated using the official HealthBench script.}
    \vspace{-2mm}
    \label{fig:pilot}
\end{figure}

\paragraph{Results}
Figure~\ref{fig:pilot} shows that training performance rises from 41.3\% to about 52\% and largely plateaus after the first epoch. In contrast, held-out performance fluctuates and shows no sustained improvement. Overall, this pilot indicates that naive SFT hits a ceiling quickly and does not generalize reliably to unseen questions.

\begin{finding}
\textbf{Generalization Failure of Naive SFT} Naive SFT quickly saturates, memorizing rubrics without achieving reliable generalization.
\end{finding}


This generalization gap motivates more generalizable supervision, especially rubric RL using instance-conditioned, scenario-aware criteria providing the necessary fine-grained rewarding.



\begin{figure*}[t]
\small
\vspace{-15pt}
\begin{AIbox}{Prompt for Drafting Instance Rubrics}
You are a rubric writer for responses to health questions in a pairwise comparison setting.

Generate 7--20 self-contained, checkable rubric items written in positive, additive form. Rubrics must be generalizable. Focus on safety, factuality, uncertainty handling, completeness, clarity.

Return only valid JSON:
\{
\ \ "rubrics\_pure": [\{"description": "..." \}, \ldots]
\}

Inputs:
question: \{...\}
ans\_a: \{...\}
ans\_b: \{...\}
preferred: \{...\}
\end{AIbox}
\vspace{-3mm}
\caption{Prompt used to draft per-instance rubrics from a clinician-labeled pairwise preference.}
\vspace{-2mm}
\label{fig:rubric_prompt_box}
\end{figure*}

\begin{figure*}[t]
\small
\begin{AIbox}{Prompt for Rubric Rewrite from Physician Feedback}
Revise the rubric based on physician feedback. Keep each item self-contained and checkable.
Return only valid JSON with the same schema.

Inputs:
question: \{...\}
rubric\_draft: \{...\}
physician\_feedback: \{..\}
\end{AIbox}
\vspace{-3mm}
\caption{Prompt used to rewrite rubrics according to physician revision and review feedback.}
\vspace{-4mm}
\label{fig:rubric_rewrite_prompt}
\end{figure*}

\section{Methodology}
\label{sec:method}

This section introduces our rubric-to-principle alignment framework, motivated by the pilot observation that naive SFT quickly saturates and generalizes poorly to novel clinical scenarios. An overview of the framework is illustrated in Figure~\ref{fig:method}. \textbf{HealthRubrics} first produces fine-grained, physician-revised rubrics from real-world medical queries and candidate responses, providing reliable preference supervision. \textbf{HealthPrinciples} then compresses recurring rubric structure into reusable, scenario-specific principles that transfer across settings. The resulting principles are used both for scalable offline training and for inference-time self-revision with rubric references. We describe HealthRubrics in Section~\ref{sec:healthrubrics} and HealthPrinciples in Section~\ref{sec:healthprinciples}.

\subsection{HealthRubrics: Granular Alignment under Physician Validation}
\label{sec:healthrubrics}

\paragraph{Medical Subset Curation.}
HealthRubrics starts from real user prompts paired with multiple candidate model answers.
We aggregate preference data from Chatbot Arena, including \texttt{human-140k}\footnote{\url{https://huggingface.co/datasets/lmarena-ai/arena-human-preference-140k}},
\texttt{human-55k}\footnote{\url{https://huggingface.co/datasets/lmarena-ai/arena-human-preference-100k}},
and \texttt{expert-5k}\footnote{\url{https://huggingface.co/datasets/lmarena-ai/arena-expert-5k}},
which were released throughout 2025 and therefore cover strong contemporary models.
We further incorporate HelpSteer3-Preference~\citep{wang2025helpsteer3preferenceopenhumanannotatedpreference}, whose prompts are sourced from user contributed ShareGPT\footnote{\url{https://huggingface.co/datasets/RyokoAI/ShareGPT52K}} and WildChat-1M~\citep{zhao2024wildchat1mchatgptinteraction}.
In total, this yields 103{,}575 queries with paired candidate responses.
To focus on healthcare, \texttt{GPT-5.1} classifies each query into medical versus non-medical categories under a fixed taxonomy.
This filtering produces 7{,}034 medical preference instances.
Additional details of the classification protocol are provided in Appendix~\ref{sec:cls}.

\paragraph{Physician Preference Consensus.}
Although the source datasets include preference labels, they mostly reflect general user judgments. To obtain clinically grounded supervision, we collect physician re-labels for each response pair, with every instance independently annotated three times. Physician agreement is higher than agreement with the original user labels, reflecting systematic differences between clinical and lay preferences. Physicians unanimously agree on 55.2\% of instances, while the rest show a two-to-one split. Consistency is moderate by kappa, with pairwise values from 0.42 to 0.51 and an overall three-annotator value of 0.47. In contrast, physician match rates with user labels are lower at 0.60 to 0.64. We therefore take majority vote over the three physician labels as the clinician consensus preference. Conditioned on this consensus and the paired responses, \texttt{GPT-5.1} drafts per-instance rubric items using the prompt in Figure~\ref{fig:rubric_prompt_box}.

\begin{table}[t]
\vspace{1mm}
\centering
\small
\begin{tblr}{
  width=\linewidth,
  colspec = {X[1.2,l,m] X[0.9,c,m] X[0.9,c,m]},
  row{1} = {bg=headerblue, font=\bfseries},
  row{even[2-Z]} = {bg=rowgray},
  cell{1}{1} = {halign=l},
}
\toprule
Revision Loop & \# Rubric Set & Proportion \\
\midrule
Loop 1 & 5{,}297 & 75.3\% \\
Loop 2 & 1{,}083 & 15.4\% \\
Loop 3 & 654     & 9.3\% \\
\bottomrule
\end{tblr}
\vspace{-1mm}
\caption{Distribution of physician revision loops required to finalize each rubric set (total $N=7{,}034$).}
\vspace{-4mm}
\label{tab:revision_loops}
\end{table}

\paragraph{Iterative Rubric Refinement.}
Draft rubrics are finalized through a two-stage physician workflow. An assigned physician first reviews the draft and provides revision suggestions by flagging incorrect items, clarifying ambiguous criteria, and identifying missing clinically relevant checks. An independent physician then audits the proposed changes and the updated rubric. A rubric set is accepted only if both physicians agree. Otherwise, it is returned for another revision loop. Table~\ref{tab:revision_loops} reports the number of loops required to finalize each rubric set over all 7034 instances. Consensus is reached in 1.34 loops on average, with a maximum of three. After consensus, \texttt{GPT-5.1} rewrites the rubric into a standardized JSON format while preserving the approved content, using prompts shown in Figure~\ref{fig:rubric_rewrite_prompt}.

\begin{table*}[t]
\vspace{-10pt}
    \centering
    {\scriptsize
    \begin{tblr}{
        width = \textwidth,
        colspec = {X[0.10,l,m] X[1.60,l,m] X[0.10,c,m]},
        row{1} = {bg=headerblue, font=\bfseries},
        row{even[2-Z]} = {bg=rowgray},
        cells = {font=\scriptsize},
    }
        \toprule
        Category & Subcategory & \#Sub~/~\#P \\
        \midrule
    
        \textbf{Urgency}
          & \texttt{\textbf{Non-Emergent}} \textit{(routine concern, no immediate safety risk)}; \texttt{\textbf{Conditionally Emergent}} \textit{(cannot rule out important risk; needs key details)}; \texttt{\textbf{Emergent}} \textit{(clear high-risk feature; immediate protective action)}
          & 3~/~13\\
        \SetCell[c=3]{l}
          \parbox{0.97\linewidth}{Example (Conditionally emergent): Communicate potential seriousness using calm, non-alarmist language; give ordered next steps: stop unsafe exposure, do low-risk actions now, avoid harms, monitor symptoms, and specify when/where to seek in-person or emergency care.}\\
    
        \midrule
        \textbf{Uncertainty}
          & \texttt{\textbf{Sufficient information}} \textit{(key info already present)}; \texttt{\textbf{Reducible uncertainty}} \textit{(missing details can be clarified in dialogue)}; \texttt{\textbf{Irreducible uncertainty}} \textit{(needs exam/measurement/testing in person)}
          & 3~/~8 \\
        \SetCell[c=3]{l}
          \parbox{0.97\linewidth}{Example (Irreducible uncertainty): For serious/complex topics, explicitly state limits (no exam/testing), avoid definitive diagnosis/prognosis/dosing, and offer a concrete low-risk way to reduce uncertainty (appropriate clinician/tests; when to seek urgent evaluation).}\\
    
        \midrule
        \textbf{Expertise}
          & \texttt{\textbf{Non-professional}} \textit{(layperson/general public)}; \texttt{\textbf{Professional}} \textit{(formal training; professional framing/terminology)}
          & 2~/~10 \\
        \SetCell[c=3]{l}
          \parbox{0.97\linewidth}{Example (Non-professional): Follow explicit user instructions on language, format, length, and scope. Stay on task and avoid unnecessary digressions.}\\
    
        \midrule
        \textbf{Task Type}
          & Emergency triage and escalation; Symptom checks and possible causes; Home care and monitoring plans; Medication and supplement safety, including interactions; Choosing tests and understanding results; Procedure and perioperative guidance; Long-term chronic care planning; Prevention, screening, and vaccines; Sexual and reproductive health counseling; Pregnancy and perinatal guidance; Mental health support and coping; Substance use harm reduction; Rehab and return-to-work or return-to-sport planning; Health education and concept explanations; Exam-style knowledge checks; Medical writing, editing, and translation; Clinical note and document drafting; Patient-facing message drafting; Care navigation and referrals; Insurance and administrative support; Non-medical requests.
          & 21~/~88 \\
        \SetCell[c=3]{l}
          \parbox{0.97\linewidth}{Example (Insurance and administrative support): Clearly distinguish what is legally required from common industry practice. Avoid overconfident claims; recommend verifying with plan documentation, clearinghouses, and applicable regulations.}\\
        \bottomrule
    \end{tblr}
    }
\vspace{-2mm}
\caption{Summary of the principle taxonomy with four parallel dimensions: Urgency, Uncertainty, Expertise, and Task Type. \texttt{\#Sub}~/~\texttt{\#P} denotes the number of subcategories and principles, respectively, and one illustrative principle is shown per category. Full definitions and induction details are provided in Appendix~\ref{sec:extract_task}.}
\vspace{-2mm}
\label{tab:labeling-schema}
\end{table*}

\begin{figure}[t]
    \centering
    \includegraphics[width=1.0\columnwidth]{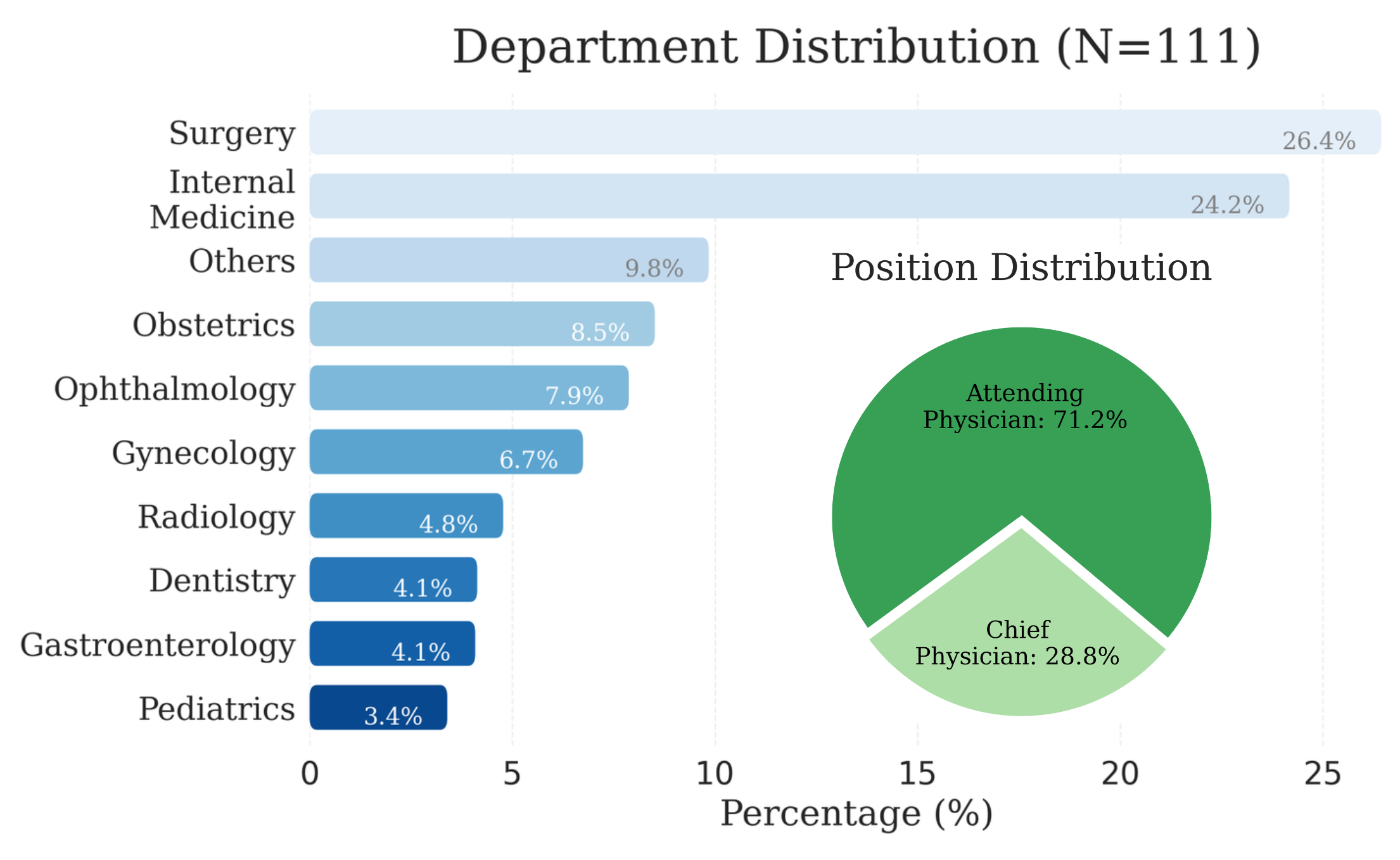}
    \vspace{-5mm}
    \caption{Distribution of physician involved.}
    \vspace{-4mm}
    \label{fig:doctor}
\end{figure}

\paragraph{Physician Cohort.}
Figure~\ref{fig:doctor} summarizes the physician cohort of 111 reviewers. Participants span a broad range of specialties, with the largest representation from surgery and internal medicine, and additional coverage across multiple departments. The cohort also spans seniority, including both attending and chief physicians. This breadth helps ensure that rubric revisions reflect practical clinical expectations across settings rather than a single specialty perspective. Overall, these activities required 632.2 person-hours at a rate of \$24 per hour, resulting in a total cost of \$15,172.80.

\subsection{HealthPrinciples: Scalable Alignment beyond Rubrics}
\label{sec:healthprinciples}

\paragraph{Motivation and Overview.}
HealthRubrics yields high-fidelity supervision, but physician validation is expensive and difficult to scale.
Moreover, many physician edits go beyond factual corrections, repeatedly reflecting scenario-dependent reasoning, safety emphasis, and communication preferences.
We therefore distill these recurring patterns into \emph{HealthPrinciples}: compact, human-readable guidance that captures what matters in a given clinical context and can be reused to generate rubrics for new questions without direct physician authorship.

\paragraph{Taxonomy Design.}
To make principles reusable and composable, we develop a top-down taxonomy with physicians that captures how clinical evaluation varies with context, as summarized in Table~\ref{tab:labeling-schema}.
The taxonomy has four dimensions: \texttt{urgency}, \texttt{uncertainty}, \texttt{user expertise}, and \texttt{task type}.
\texttt{Urgency} has three levels: non-emergent, conditionally emergent, and emergent.
\texttt{Uncertainty} is grouped into sufficient information, reducible uncertainty, and irreducible uncertainty.
\texttt{User expertise} is grouped into non-professional and professional.
For \texttt{task type}, we use \texttt{GPT-5.1} to extract a primary task and an optional secondary task without a predefined label set, then cluster the extracted tasks and consolidate them with physicians into 21 task types. Definitions of urgency, uncertainty, and expertise appear in Appendix~\ref{sec:taxonomy_urgency}, ~\ref{sec:taxonomy_urgency} ~and~\ref{sec:taxonomy_expertise}.
Task Tyoe's definitions and the induction procedure appear in Appendix~\ref{sec:taxonomy_tasktype}.

\paragraph{From Rubrics to Principles.}
After finalizing the taxonomy, we map each rubric to one or more subcategories. Within each subcategory, we cluster rubrics by semantic similarity and derive candidate principles that summarize recurring patterns. We then refine these principles with physicians until consensus is reached, ensuring clarity and clinical faithfulness. This yields 119 principles, with one to five principles per subcategory. Examples appear in Table~\ref{tab:labeling-schema}, and details are provided in Appendix~\ref{sec:principle_extraction}.

\begin{figure*}[t]
\small
\begin{AIbox}{Prompt for Rubric Generation from Principles}
You are generating a rubric to evaluate answers to a medical question. Given:\\
- question: the user query (may include multi-turn context) \\
- principles: scenario-specific principles (may be incomplete or partially mismatched) \\
\textbf{Goal}: Convert principles into 7--20 concrete, checkable rubric items for grading one answer.\\
Rubrics must be observable behaviors that are easy to verify directly from the text.\\
Return ONLY valid JSON. Inputs: \{...\} question: \{...\} principles: \{...\}

\end{AIbox}
\vspace{-2mm}
\caption{Prompt for generating per-question rubrics conditioned on extracted HealthPrinciples.}
\vspace{-2mm}
\label{fig:rubric_from_principles_prompt}
\end{figure*}

\paragraph{Principle-conditioned Rubric Generation.}
HealthPrinciples enables scalable, rubric-grounded supervision for unseen queries.
We collect 16{,}872 additional medical questions from UltraMedical-Preference~\citep{zhang2024ultramedicalbuildingspecializedgeneralists}, using the ChatDoctor~\citep{li2023chatdoctormedicalchatmodel} and MedQuAD~\citep{BenAbacha-BMC-2019} subsets.
For each question, \texttt{GPT-5.1} assigns taxonomy labels for urgency, uncertainty, and user expertise, and predicts a primary task with an optional secondary task, using the prompts in Appendix~\ref{sec:question_cls}.
The corresponding principles are then extracted and converted into question-specific, scorable rubric items.
Each question retrieves 22.9 principles on average.
Figure~\ref{fig:rubric_from_principles_prompt} shows the prompt for converting principles into rubric items.

\paragraph{Inference-time Rubric Guidance.}
We package the workflow as a reusable inference-time tool.
Given a question, optionally with dialogue context or a draft answer, the tool classifies the scenario, extracts the matched principles, and generates context-specific rubric items.
These rubric references can guide self-revision and enable scenario-aware improvements beyond direct imitation.
Implementation details and prompts are provided in Appendix~\ref{sec:principle_tool}.

\section{Experiments}

\subsection{Experimental Setup}
\label{sec:exp_setup}

\paragraph{Evaluation Benchmarks.}
We evaluate on three complementary benchmarks. HealthBench~\citep{arora2025healthbench} is an open-source benchmark of 5,000 multi-turn healthcare conversations involving both lay users and clinicians, designed to assess clinical usefulness and safety. LLMEval-Med~\citep{zhang2025llmeval} comprises 2,996 questions spanning five core medical areas, drawn from real-world EHRs and expert-crafted clinical scenarios; we report results on its Medical Language Understanding, Medical Reasoning, and Medical Safety and Ethics subsets. Arena-Hard-v2~\citep{li2024crowdsourceddatahighqualitybenchmarks} includes 500 challenging real-world prompts curated by BenchBuilder and is widely used as an automatic open-ended evaluation proxy, showing strong correlation with Chatbot Arena.

\paragraph{Training Data and Rubric Variants.}
As described in Section~\ref{sec:method}, we sample 7{,}034 medical questions from the medical portions of Chatbot Arena and HelpSteer3-Preference~\citep{wang2025helpsteer3preferenceopenhumanannotatedpreference} and construct three rubric supervision variants for each question. As described in Section~\ref{sec:healthrubrics}, \textbf{Draft Rubrics} are generated automatically and \textbf{Doctor Rubrics} are revised by physicians, while Section~\ref{sec:healthprinciples} introduces \textbf{Principle Rubrics} which extract matched \texttt{HealthPrinciples} and convert them into question-specific rubric items. To leverage the scalability of principles, we additionally collect 16{,}872 medical questions from UltraMedical-Preference~\citep{zhang2024ultramedicalbuildingspecializedgeneralists} using the ChatDoctor~\citep{li2023chatdoctormedicalchatmodel} and MedQuAD~\citep{BenAbacha-BMC-2019} subsets and generate Principle Rubrics for all of them. All training questions are real user medical queries from online platforms, which we intentionally choose to minimize human-introduced bias and better reflect real-world user needs.

\paragraph{Baseline Models.}
We compare against ten strong baselines, including open models DeepSeek-R1~\citep{deepseekai2026deepseekr1incentivizingreasoningcapability}, Qwen3-235B-Instruct, and Qwen3-32B~\citep{yang2025qwen3technicalreport}, proprietary frontier models \texttt{o3}, Claude-3.7-Sonnet, Gemini-2.5-Pro~\citep{comanici2025gemini25pushingfrontier}, and GPT-4.1~\citep{openai2024gpt4technicalreport}, and medical-domain models Baichuan-M2~\citep{dou2025baichuan} and HuatuoGPT-o1~\citep{chen2024huatuogpt}.

\begin{table*}[t]
\centering
\footnotesize

\begin{tblr}{
  width=\linewidth,
  colspec = {X[2.8,l,m] X[0.85,c,m] X[0.85,c,m] X[0.95,c,m] X[1.10,c,m] X[1.05,c,m] X[0.85,c,m] X[1.10,c,m]},
  rowsep=1.5pt,
  row{1-2} = {bg=headerblue, font=\bfseries},
  row{even[3-Z]} = {bg=rowgray},
  cell{1}{1} = {halign=l, valign=m},
  vline{2,4,7} = {0.6pt, gray6},
  hline{1,Z} = {1pt},
  hline{3}   = {0.8pt, gray6},
  hline{4,7,12,15} = {0.6pt, gray6},
}
\SetCell[r=2]{l} Model
& \SetCell[c=2]{c} HealthBench
&  & \SetCell[c=3]{c} LLMEval-Med
&  &  & \SetCell[c=2]{c} Arena-Hard-v2
& \\
& Hard & Overall
& Reasoning & Safety \& Ethics & Under- standing
& Hard & Creative Writing \\

\SetCell[c=8]{c, font=\bfseries} Open-source LLMs \\
Deepseek-R1 & 15.1 & 47.4 & 63.4 & 69.6 & 69.6 & 56.8 & 77.0 \\
Qwen3-235B-Instruct & 16.2 & 50.0 & 65.7 & 76.2 & 61.5 & 46.7 & 73.5 \\
Qwen3-32B & 12.0 & 46.1 & 59.1 & 62.3 & 59.3 & 44.5 & 53.3 \\

\SetCell[c=8]{c, font=\bfseries} Closed-source LLMs \\
o3 & 31.6 & 59.8 & 63.8 & 66.9 & 65.8 & 85.9 & 88.8 \\
Claude-3.7-Sonnet & 15.0 & 34.6 & 57.8 & 75.2 & 54.0 & 59.8 & 63.9 \\
Gemini-2.5-Pro & 18.5 & 52.0 & 73.5 & 72.5 & 60.1 & 79.0 & 90.8 \\
GPT-4.1 & 16.0 & 47.9 & 60.4 & 57.3 & 58.8 & 50.0 & 61.5 \\

\SetCell[c=8]{c, font=\bfseries} Specialized LLMs \\
Baichuan-M2-32B & 34.7 & 60.1 & 64.2 & 63.8 & 64.2 & 45.8 & 69.2 \\
HuatuoGPT-o1-72B & 18.2 & 47.9 & 56.9 & 56.3 & 49.5 & 43.2 & 55.3 \\

\SetCell[c=8]{c, font=\bfseries} Our Method \\
Qwen3-4B-Instruct & 5.2 & 40.6 & 39.2 & 66.1 & 49.8 & 15.0 & 13.2 \\
\SetCell{l}\hspace*{1em}+~\textit{Draft Rubrics}
  & \scoredelta{21.2}{+16.0} & \scoredelta{46.9}{+6.3} & \scoredelta{39.2}{+0.0}
  & \scoredelta{85.3}{+19.2} & \scoredelta{52.6}{+2.8} & \scoredelta{34.9}{+19.9} & \scoredelta{50.5}{+37.3} \\
\SetCell{l}\hspace*{1em}+~\textit{Doctor Rubrics}
  & \scoredelta{22.9}{+17.7} & \scoredelta{51.0}{+10.4} & \scoredelta{46.1}{+6.9}
  & \scoredelta{81.7}{+15.6} & \scoredelta{55.9}{+6.1} & \scoredelta{39.7}{+24.7} & \scoredelta{55.3}{+42.1} \\
\SetCell{l}\hspace*{1em}+~\textit{Principle Rubrics}
  & \scoredelta{24.4}{+19.2} & \scoredelta{51.1}{+10.5} & \scoredelta{42.2}{+3.0}
  & \scoredelta{80.7}{+14.6} & \scoredelta{52.6}{+2.8} & \scoredelta{37.0}{+22.0} & \scoredelta{51.0}{+37.8} \\
\SetCell{l}\hspace*{1em}+~\textit{More Query Rubrics}
  & \scoredelta{27.2}{+22.0} & \scoredelta{52.9}{+12.3} & \scoredelta{44.7}{+5.5}
  & \scoredelta{87.2}{+21.1} & \scoredelta{55.4}{+5.6} & \scoredelta{41.2}{+26.2} & \scoredelta{61.3}{+48.1} \\
Qwen3-30B-A3B-Instruct & 15.0 & 46.8 & 54.9 & 67.0 & 52.1 & 33.9 & 34.9 \\
\SetCell{l}\hspace*{1em}+~\textit{More Query Rubrics}
  & \scoredelta{33.4}{+18.4} & \scoredelta{59.5}{+12.7} & \scoredelta{59.8}{+4.9}
  & \scoredelta{79.8}{+12.8} & \scoredelta{57.8}{+5.7} & \scoredelta{74.6}{+40.7} & \scoredelta{79.4}{+44.5} \\
\end{tblr}
\caption{Model results on HealthBench, LLMEval-Med, and Arena-Hard-v2. \textit{Draft Rubrics} are auto-generated rubrics; \textit{Doctor Rubrics} are physician-revised; \textit{Principle Rubrics} are derived from matched \texttt{HealthPrinciples}; \textit{More Query Rubrics} scales supervision with additional medical queries.}
\vspace{-4mm}
\label{tab:model_results}
\end{table*}

\paragraph{RL Rraining.}
We perform reinforcement learning with GRPO~\citep{shao2024deepseekmathpushinglimitsmathematical} implemented in the \texttt{verl} framework~\citep{Sheng_2025}. Unless otherwise noted, we use 8 rollouts per prompt, a learning rate of $10^{-6}$, and a batch size of 64, training on four H200 nodes. For all rubric-supervised variants, rubric-based scoring uses a fixed judge model, Qwen3-32B. We stop training when performance saturates, which typically occurs within 20 optimization steps in our main runs.

\subsection{Main Results}
\label{sec:exp_results}

Table~\ref{tab:model_results} summarizes the main results on HealthBench, LLMEval-Med, and Arena-Hard-v2.

\paragraph{Expert-verfied Rubrics}
Rubric conditioned RL remains effective under automatically generated supervision. Training with \textbf{Draft Rubrics} yields strong gains over models across benchmarks, consistent with the pilot results in Section~\ref{sec:pilot} and showing that rubric based rewards provide a learnable and generalizable signal.
\textbf{Doctor Rubrics} further improve over Draft Rubrics, with the largest gains on HealthBench in both Hard and Overall settings and on Arena-Hard-v2. This gap suggests that physician edits remove ambiguity and fix mis specified criteria, producing higher fidelity rewards and better policy updates. Doctor supervision also makes safety critical behaviors such as triage and escalation more consistent and reduces reward hacking driven by vague rubric items, which together explains the stronger overall gains.

\paragraph{From Rubrics to Principles}
\textbf{Principle Rubrics} are competitive with physician edited rubrics even without additional scaling. Trained on the same 7{,}034 questions, they match and sometimes slightly exceed Doctor Rubrics, suggesting that the extracted \emph{HealthPrinciples} capture transferable evaluation structure rather than dataset specific artifacts.
A key factor is coverage. Principle retrieval pools recurring requirements across semantically similar questions, so the rubric for a given query typically spans more facets such as risk assessment, uncertainty handling, and actionable next steps than a purely question local rubric.
Scaling principles to more questions further improves results. \textbf{More Query Rubrics} add principle generated rubrics for 16{,}872 real user medical queries and deliver broad gains on HealthBench and Arena-Hard-v2. This pattern is consistent with greater robustness from diversified and scenario matched supervision.

\begin{figure}[t]
    \centering
    \includegraphics[width=0.96\columnwidth]{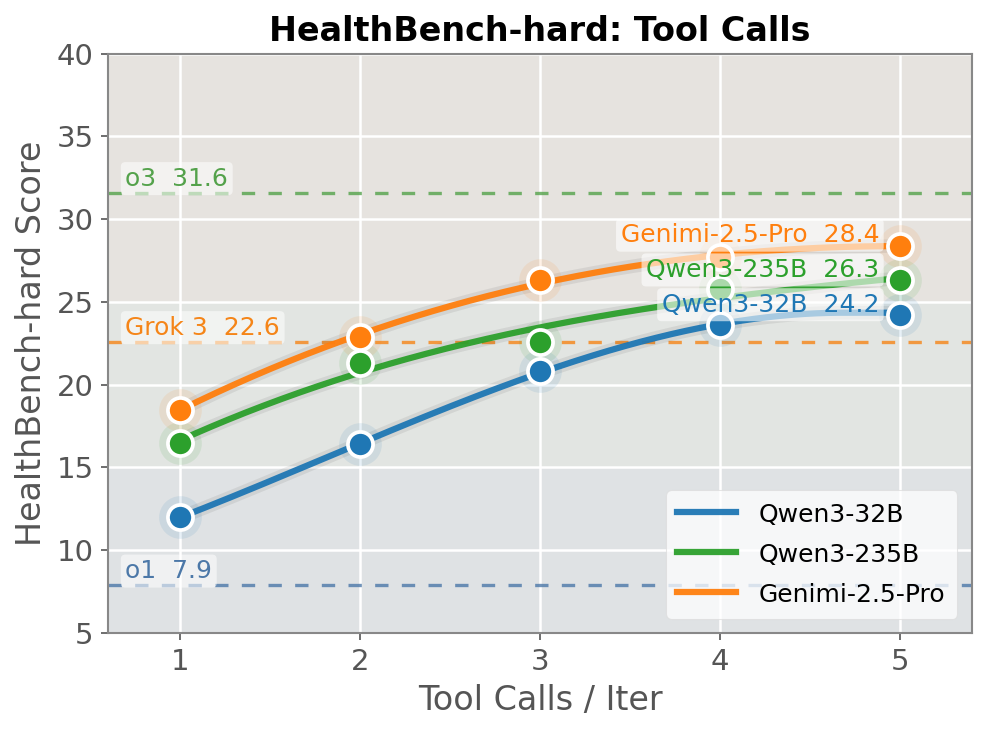}
    \vspace{-2mm}
    \caption{Tool-call Scaling on HealthBench-hard.}
    \vspace{-4mm}
    \label{fig:tool_calls}
\end{figure}

\paragraph{From Medical Alignment to General Usefulness}
Although training optimizes medical rubric rewards, gains transfer to the open ended Arena-Hard-v2 setting. This suggests that rubric grounding strengthens general instruction following behaviors such as tracking user intent, organizing actionable guidance, and communicating limitations appropriately.
In contrast, explicitly reasoning heavy metrics such as LLMEval-Med \emph{Reasoning} change little, indicating that rubric aligned training mainly improves \emph{helpfulness and safety} rather than problem solving depth. A promising next step is to combine rubric based RL with objectives that more directly train multi step reasoning, with the goal of improving both alignment and reasoning.

\paragraph{Inference-time Scaling}
\label{sec:exp_tool_scaling}
Figure~\ref{fig:tool_calls} shows that allowing multiple inference-time calls to our rubric-guidance tool consistently improves HealthBench-hard performance across backbones, even without any specific training, indicating that extracted principles and generated rubrics provide actionable revision signals at test time.
Performance increases with more iterations but gradually saturates after a few calls, suggesting diminishing returns once major rubric mismatches are corrected and the remaining errors are harder to fix through further rubric feedback alone.

\section{Analysis}
\label{sec:analysis}

\paragraph{Question Scaling with Fixed Budget}
Under a fixed compute budget, this subsection evaluates how performance scales with question coverage.
Following Section~\ref{sec:exp_setup}, we randomly subsample 1k, 2.5k, 5k, 10k, and 20k questions from the full training pool.
Training FLOPs are held constant by matching each run to training on 20k questions for two epochs, so smaller datasets are trained for proportionally more epochs, such as four for 10k and eight for 5k. For each setting, we evaluate the best checkpoint on HealthBench-hard.

\begin{figure}[t]
    \centering
    \includegraphics[width=0.96\columnwidth]{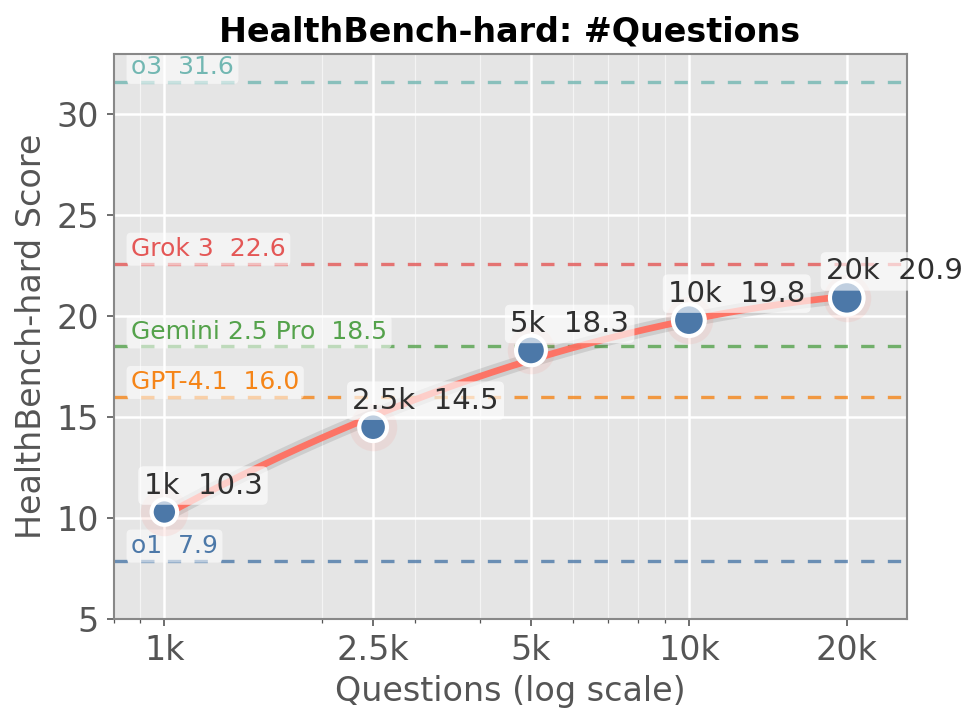}
    \vspace{-2mm}
    \caption{Question scaling on HealthBench-hard under a fixed training-FLOPs budget with Qwen3-4B-Instruct.}
    \vspace{-4mm}
    \label{fig:question_scale}
\end{figure}

Figure~\ref{fig:question_scale} shows a monotonic improvement as the number of distinct questions increases.
The gains are largest from 1k to 5k and remain positive up to 20k, with diminishing returns at larger scales.
This indicates that rubric-based RL benefits more from supervision diversity than from additional epochs on a narrow prompt set, likely because broader coverage better spans clinical intents, risk profiles, and failure modes.
Overall, scaling is most effective by increasing the number of training questions.

\paragraph{Rubrics Scoring Model Choice}
To select an efficient and accurate rubric-satisfaction scorer, we randomly sample 1,000 HealthBench questions and evaluate model answers from both training and evaluation, yielding 11,446 rubric judgments. Treating the HealthBench protocol with GPT-4.1 as the reference, we measure Qwen3 scoring accuracy, which increases with model scale: Qwen3-4B, 14B, 32B, and 235B achieve 76.4\%, 80.2\%, 87.6\%, and 87.9\%, respectively. Given the negligible gain from 32B to 235B, we use Qwen3-32B as our default scorer for the best accuracy–efficiency trade-off.


\section{Conclusion}
In this work, we addressed the challenge of aligning medical language models with fine-grained standard.
We introduced a robust framework anchored by \textbf{HealthRubrics}, a physician-verified dataset, and \textbf{HealthPrinciples}, a reusable taxonomy that distills expert consensus for scalable supervision. 
Through three progressive strategies: learning on validated data, scaling via principle-synthesized rubrics, and inference-time guidance, we demonstrated that such an alignment enables models to surpass frontier proprietary systems. 
These findings suggest that incorporating structured clinical logic is as effective as scaling model parameters for specialized tasks. 
By releasing our data, principles, and tool, we provide a practical, resource-efficient foundation to accelerate future research into safe and reliable healthcare AI.
\clearpage

\section*{Limitations}
Despite the overall gains, we identify two key limitations. First, the proposed method does not yield a substantial improvement in intrinsic reasoning in our experiments. Although it helps the model produce better answers in some cases, the gains on multi-step reasoning tasks are not consistently significant. This suggests that the method primarily improves response quality by leveraging external signals more effectively, rather than fundamentally strengthening internal reasoning, and leaves room for further refinement through more targeted supervision and tighter integration between reasoning and tool use.

Second, as a inference-time scaling strategy, the benefits saturate quickly. Increasing tool usage or sampling beyond a certain point produces diminishing returns, indicating that simply scaling inference-time tool calls is unlikely to deliver sustained improvements. A promising direction is to train a more agentic model that can plan, decide when to invoke tools, and coordinate tool use more effectively, enabling better credit assignment and more efficient inference-time computation.

\newpage

\bibliography{custom}

\clearpage
\appendix
\section{SFT Data Generation Prompt}
\label{app:sftprompt}

To reduce brittleness from imitating a single reference completion, we expand each training question into multiple rubric-aware targets. For each question, we provide \texttt{GPT-5.1} with the original question, the official HealthBench rubrics, a current model draft response, and a clinician-written ideal completion. The draft and ideal serve as complementary anchors, highlighting common failure modes and desired coverage, while the rubrics define the scoring contract the generated answer should satisfy. Using the template in Figure~\ref{fig:sft_prompt}, we sample three independent answers per question to create stylistic and structural variants that aim to meet the same rubric requirements. This produces a 9,000-instance SFT dataset from 3,000 training questions, used in the pilot SFT experiments in Section~\ref{sec:pilot}.

\begin{figure*}[t]
\begin{AIbox}{Prompt for Rubric-Aware Answer Generation with GPT-5.1}
You are a careful clinician-style medical assistant. You will write an answer to a patient-facing health question.\\
Your goal is to maximize the HealthBench rubric score.\\
\\
Task\\
Given a question and its rubrics, produce a high-quality answer that satisfies as many rubric items as possible.\\
You will also see a model draft answer and a clinician ideal completion. Use them as reference only.\\
Do not copy sentences verbatim. Write a fresh answer.\\
\\
Hard requirements\\
- Write in ENGLISH.\\
- Follow the rubrics closely and include key safety guidance and triage when applicable.\\
- Do not invent facts. If needed, ask clarifying questions.\\
- Be specific and actionable; use short paragraphs and bullet points when helpful.\\
- Avoid policy or rubric talk. Do not mention rubrics, scoring, evaluation, or the prompt.\\
- Do not mention the draft or the ideal completion.\\
- Output only the final answer text.\\
\\
Inputs\\
Question\\
\{question\}\\
\\
Rubrics\\
\{rubrics\}\\
\\
Model draft answer\\
\{draft\}\\
\\
Clinician ideal completion\\
\{ideal\}\\
\\
Output\\
Write one final answer that best satisfies the rubrics.
\end{AIbox}
\caption{Prompt template for generating rubric-aware SFT targets with \texttt{GPT-5.1}. We generate three answers per training question to increase coverage and reduce overfitting to a single phrasing.}
\label{fig:sft_prompt}
\end{figure*}

\section{Medical Query Classification for HealthRubrics}
\label{sec:cls}

To extract a high precision medical subset from the pooled preference corpora we use \texttt{GPT 5.1} to classify each conversation as medical related or not using a fixed guideline
A conversation is labeled \textsc{Medical} if a responsible response would require clinical or biomedical knowledge including diagnosis treatment medication use prognosis safety risks or interpretation of medical information
Conversations that are mainly administrative or only loosely health related without clinical reasoning are labeled \textsc{Non medical}
We use deterministic decoding for consistency and manually spot check a random sample to confirm precision
The prompt is shown in Figure~\ref{fig:medical_cls_prompt}

\begin{figure*}[ht]
\begin{AIbox}{Prompt for Medical Query Classification full}
You are a medical text classifier Given the following conversation determine if it is medical related. You must only output a single JSON object. The JSON must have exactly one key \"is\_medical\". The value must be either true or false in lowercase. Do not output any explanation or additional text.Conversation: \{conversation\}
\end{AIbox}
\vspace{-4mm}
\caption{Prompt used to filter medical related conversations prior to rubric construction}
\label{fig:medical_cls_prompt}
\end{figure*}

\section{Subcategory Definitions and Task Type Induction}
\label{sec:extract_task}

This appendix provides the full definitions for the HealthPrinciples taxonomy used in Section~\ref{sec:healthprinciples}.
Table~\ref{tab:labeling-schema} in the main text presents a compact view; here we detail every subcategory and describe how the task-type inventory was induced from data with physician input.

\subsection{Urgency}
\label{sec:taxonomy_urgency}

\begin{itemize}[leftmargin=1.2em, itemsep=2pt, topsep=2pt]
    \item \textbf{Non-emergent.} No immediate safety threat is suggested; the query concerns routine health issues or general information and typically does not require time-sensitive action.
    \item \textbf{Conditionally emergent.} Time-sensitive risk cannot be ruled out from the given information; urgency depends on missing key details. The response should ask targeted clarifying questions and provide conditional escalation guidance with clear red flags.
    \item \textbf{Emergent.} The query indicates high-risk features or an immediate safety threat and warrants prompt protective action and escalation guidance.
\end{itemize}

\subsection{Uncertainty}
\label{sec:taxonomy_uncertainty}

\begin{itemize}
    \item \textbf{Sufficient information.} The question includes the key details needed to give safe and effective guidance; no essential clarifications are required to proceed.
    \item \textbf{Reducible uncertainty.} Important details are missing or ambiguous but can be obtained through follow-up questions in dialogue.
    \item \textbf{Irreducible uncertainty.} The uncertainty cannot be resolved remotely and requires examination, objective measurement, diagnostic testing, or in-person professional evaluation.
\end{itemize}

\subsection{User expertise}
\label{sec:taxonomy_expertise}

\begin{itemize}
    \item \textbf{Non-professional.} The user is a layperson or general public; responses should prioritize plain language and explicit actionability.
    \item \textbf{Professional.} The user indicates relevant formal training or uses professional framing and terminology; responses may use technical language while maintaining safety.
\end{itemize}

\subsection{Task Type Inventory}
\label{sec:taxonomy_tasktype}

Each query is assigned a primary task and an optional secondary task.
Task types are designed to be readable labels aligned with Table~\ref{tab:labeling-schema}; we avoid underscored identifiers in the paper text.

\paragraph{Task type induction.}
We do not pre-specify a task taxonomy.
Instead, we first ask \texttt{GPT-5.1} to freely summarize the \emph{primary} and \emph{secondary} task implied by each query in short natural language.
We then cluster these task summaries using \texttt{GPT-5.1} to propose multiple candidate clusterings at different granularities.
Finally, physicians review the candidates, reconcile ambiguous boundaries, and select a 21-category inventory that best matches clinical practice and dataset coverage.
During this process, physicians additionally introduced the task family \emph{Procedure and perioperative guidance}, which was under-represented in early model-generated clusterings but frequently appeared in real patient questions and clinician feedback.

\paragraph{Free-form task extraction prompt.}
Figure~\ref{fig:task_extract_prompt} shows the template used to extract primary and secondary tasks before clustering.

\begin{figure*}[t]
\vspace{-2mm}
\begin{AIbox}{Prompt for Free-form Task Extraction}
You are analyzing a medical user question.\\
Identify:\\
1. the primary task the assistant is being asked to perform \\
2. secondary tasks if present, otherwise None \\
Write tasks as short natural-language phrases, not as labels from a fixed list. \\
Do not give medical advice. Only describe the task(s). \\
question: \{question\}
\end{AIbox}
\vspace{-2mm}
\caption{Prompt used to extract free-form descriptions of tasks types prior to inducing the task taxonomy.}
\vspace{-2mm}
\label{fig:task_extract_prompt}
\end{figure*}

\paragraph{Final task families and definitions.}
Below are the 21 task families used in this work.

\begin{itemize}
    \item \textbf{Emergency triage and escalation.} Assess whether an emergency or time-sensitive risk is present and whether immediate escalation is required, and specify clear red flags and time windows for seeking care. Response considerations: prioritize high-yield red flags and explicit timelines; avoid undue alarm or false reassurance; default to conservative in-person evaluation when key information is missing.
    \item \textbf{Symptom assessment and differential considerations.} Provide an initial assessment based on symptoms and history, outline plausible etiologies, and identify missing information that would change risk or management. Response considerations: distinguish common versus high-risk causes and state rationale; ask targeted clarifying questions; avoid presenting possibilities as a definitive diagnosis.
    \item \textbf{Self-care, monitoring, and follow-up.} Offer home-care and self-management guidance and define thresholds and time windows for follow-up, reassessment, strategy changes, or escalation. Response considerations: provide actionable steps and measurable targets; specify stop rules and escalation triggers; avoid high-risk procedures.
    \item \textbf{Medication and supplement safety.} Provide general guidance on medication or supplement use with emphasis on interactions, adverse effects, contraindications, and when professional confirmation is needed. Response considerations: account for comorbidities, pregnancy, renal or hepatic impairment, and concomitant drugs; avoid prescribing-level dosing or substitution directives.
    \item \textbf{Diagnostics and test interpretation.} Discuss test selection and interpretation, address limitations, and provide results-contingent next steps such as repeat testing, additional workup, referral, or observation. Response considerations: explain reference ranges and false positives or negatives; avoid definitive conclusions from isolated values.
    \item \textbf{Procedure and perioperative guidance.} Provide peri-procedural and perioperative guidance, including preparation, recovery, complication recognition, and relevant follow-up milestones. Response considerations: emphasize warning signs and return precautions; avoid clinician-specific instructions that require the procedural or anesthesia team; reinforce individualized discharge instructions.
    \item \textbf{Chronic disease management.} Describe a long-term management framework including goals, monitoring cadence, lifestyle and pharmacologic strategies, complication prevention, and self-management. Response considerations: present a structured plan; emphasize adherence and longitudinal follow-up; avoid one-size-fits-all targets or unsupervised medication changes.
    \item \textbf{Prevention, screening, and vaccines.} Provide prevention and risk-reduction guidance including vaccination, screening, health behaviors, and post-exposure actions. Response considerations: specify eligibility, contraindications, and time windows; note guideline and local variation; avoid absolute guarantees.
    \item \textbf{Sexual and reproductive health counseling.} Provide education and counseling on sexual health, contraception, sexual function, and common concerns. Response considerations: use nonjudgmental language; address STI risk and testing; include escalation criteria for severe pain, heavy bleeding, and safety concerns.
    \item \textbf{Pregnancy and perinatal care.} Address pregnancy and perinatal health issues including symptom evaluation, antenatal pathways, medication and lifestyle considerations, and postpartum recovery. Response considerations: use conservative safety thresholds; flag medications needing obstetric confirmation; state key triggers such as bleeding and hypertensive symptoms.
    \item \textbf{Mental health support.} Provide mental health information and support, risk identification, and guidance on when to seek professional or emergency help. Response considerations: prioritize self-harm or violence risk screening and crisis pathways; avoid substituting for professional diagnosis.
    \item \textbf{Substance use harm reduction.} Provide counseling using a harm-reduction approach including withdrawal and relapse considerations and referral indications. Response considerations: avoid actionable details that facilitate misuse; highlight overdose and withdrawal warning signs; maintain supportive language.
    \item \textbf{Rehabilitation and return to activity.} Support rehabilitation and functional recovery with graded plans, restrictions, milestones, and reassessment points for return to sport, work, or school. Response considerations: define stop rules; provide measurable progression; specify contraindications.
    \item \textbf{Health education and general explanations.} Provide accessible explanations of biomedical concepts and terminology. Response considerations: use layered explanations; communicate uncertainty; avoid drifting into individualized clinical advice.
    \item \textbf{Knowledge check and exam preparation.} Support clinician- or exam-oriented questions with structured takeaways and workflows. Response considerations: note regional guideline variability; avoid fabricating guideline details when uncertain.
    \item \textbf{Medical writing, editing, and translation.} Support medical or scientific writing and translation with terminology consistency and appropriate register. Response considerations: preserve meaning without inventing results or citations; flag ambiguous source phrasing.
    \item \textbf{Clinical documentation drafting.} Draft professional clinical records such as SOAP notes, referral letters, or discharge summaries. Response considerations: prioritize objective, time-anchored facts and traceable assessment and plan; protect privacy.
    \item \textbf{Patient-facing communication and wording.} Develop patient- or public-facing materials emphasizing clarity, tone, and actionable guidance. Response considerations: use plain language; communicate uncertainty; avoid absolutes.
    \item \textbf{Care navigation and referrals.} Provide guidance on which service to contact, how to prepare for visits, and follow-up planning. Response considerations: provide concrete checklists; avoid presenting navigation guidance as diagnostic conclusions.
    \item \textbf{Insurance and administrative processes.} Address insurance, certification, and compliance processes where the deliverable is procedural guidance rather than clinical decision-making. Response considerations: defer to institutional requirements; avoid facilitating misrepresentation; flag privacy and compliance risks.
    \item \textbf{Non-medical.} Queries that do not require clinical reasoning. Response considerations: respond without unnecessary healthcare framing; if a hidden health concern emerges, suggest reframing into an appropriate medical task type.
\end{itemize}

\section{Principle Extraction from Rubrics}
\label{sec:principle_extraction}

This appendix describes how we derive \emph{HealthPrinciples} from physician-validated rubric items.
The extraction pipeline has two stages: (i) Taxonomy-based routing of rubric sets into subcategories and (ii) Iterative clustering and compression within each subcategory to produce candidate principles for physician refinement.

\subsection{Stage I: Mapping rubrics to taxonomy subcategories}
\label{sec:principle_extraction_routing}

Each validated rubric set is assigned to one or more taxonomy subcategories.
We formulate this as a multi-label classification problem over the full set of subcategories across the four taxonomy axes.
We use \texttt{GPT-5.1} with a constrained label set and a strict JSON output format to ensure routable, consistent outputs.

\begin{figure*}[t]
\begin{AIbox}{Prompt for Mapping a Rubric Set to Taxonomy Subcategories (template)}
You are assigning taxonomy labels to a rubric set for medical answer evaluation.\\
\\
You will be given:\\
- question: the original user question\\
- rubrics: a list of rubric items for evaluating answers to the question\\
- taxonomy: the full set of allowed subcategory labels for four axes:\\
  Urgency (\{subcategories\}), Uncertainty (\{subcategories\}) \\
  Expertise (\{subcategories\}), Task Type (\{21 families\})\\
\\
Task:\\
Select ALL subcategories that the rubrics reflect.\\
This is multi-label: a rubric set may map to multiple task types and may include guidance relevant to more than one axis.\\
Choose only from the provided label lists.\\
If uncertain, choose the minimal set that still covers the rubrics.\\
\\
Output format (STRICT JSON only):\\
\{\\
  "urgency": ["A1" | "A2" | "A3"],\\
  "uncertainty": ["B1" | "B2" | "B3"],\\
  "expertise": ["C1" | "C2"],\\
  "task\_type": ["<Task Family Name>", ...],\\
  "rationale": "one short sentence"\\
\}\\
\\
Inputs:\\
question:\\
\{question\}\\
\\
rubrics:\\
\{rubrics\}\\
\\
taxonomy (allowed labels):\\
\{taxonomy\}
\end{AIbox}
\caption{Prompt used to route each rubric set into one or more taxonomy subcategories prior to clustering.}
\label{fig:rubric_to_subcategory_prompt}
\end{figure*}

\subsection{Stage II: Iterative clustering and compression within each subcategory}
\label{sec:principle_extraction_clustering}

After routing, each subcategory contains a corpus of rubric items.
To summarize recurring patterns at scale, we apply semantic clustering with iterative compression.
We compute rubric-item embeddings using \texttt{Qwen3-Embedding-8B} (L2-normalized), cluster them with \texttt{MiniBatchKMeans}, and select representative items by cosine similarity to each cluster centroid.
We then ask \texttt{GPT-5.1} to summarize each cluster into a single representative sentence.
The resulting candidates are reclustered and re-summarized iteratively until a compact set is produced for physician review.

\paragraph{Compression schedule.}
We use a fixed compression ratio of 60: each iteration compresses roughly 60 candidate items into 1 representative candidate.
We run up to three iterations, enabling hierarchical coverage of up to $60 \times 60 \times 60$ items.
When the remaining candidate count drops below 100, we directly summarize them into 5 candidates to form a manageable set for physician refinement.

\paragraph{Summarization constraints.}
All summarization prompts are in English and require strict JSON outputs, as shown in Figure~\ref{fig:cluster_summarization_prompt}.
Candidates are single-sentence, observable, and checkable criteria (i.e., rubric-like statements rather than abstract advice).
Physicians subsequently edit and consolidate these candidates into the final HealthPrinciples inventory.

\begin{figure*}[t]
\begin{AIbox}{Prompt for Cluster Summarization into Representative Rubric Candidates (template)}
You are an expert rubric editor for evaluating responses to health/medical questions.\\
\\
Task:\\
Rewrite the input rubric sentences into exactly \{k\} representative rubric sentences in ENGLISH.\\
\\
Hard requirements:\\
- Output EXACTLY \{k\} items.\\
- Return ONLY a JSON array of exactly \{k\} strings (no extra text).\\
- Each item MUST be a single sentence, self-contained, specific, and checkable by a non-expert.\\
- Each item MUST capture a distinct recurring requirement; avoid duplicates/overlap.\\
- Each item MUST be written as an observable criterion (e.g., ``Mentions X'', ``Advises Y when Z''), not an abstract principle (avoid ``be empathetic/clear/accurate'' unless tied to concrete observable elements).\\
- Keep each sentence short (preferably $\le$ 25 words); remove hedging and avoid multi-clause lists unless essential.\\
- Do NOT mention ``the list'', ``above'', ``these items'', files, prompts, or any process.\\
- Do NOT introduce new medical requirements not present in the input.\\
\\
Input rubrics:\\
\{joined\}\\
\\
Output format (STRICT):\\
Return ONLY a JSON array of exactly \{k\} strings.
\end{AIbox}
\caption{Prompt used to summarize each rubric-item cluster into exactly $k$ representative, rubric-like candidates under strict constraints.}
\label{fig:cluster_summarization_prompt}
\end{figure*}

\section{Question Classification Prompts}
\label{sec:question_cls}

This appendix presents the prompts used for taxonomy prediction and rubric construction.
Given a user question (possibly multi-turn), \texttt{GPT-5.1} predicts taxonomy labels for \texttt{urgency}, \texttt{uncertainty}, and \texttt{expertise}, and assigns exactly one primary task family with up to two optional secondary task families.
All predictions are restricted to the predefined label set (Appendix~\ref{sec:extract_task}) and returned as strict JSON for downstream retrieval.
The question-classification prompt is shown in Figure~\ref{fig:question_cls_prompt}, and the rubric-generation prompt is shown in Figure~\ref{fig:principle_tool_rubric_prompt}.

\begin{figure*}[t]
\begin{AIbox}{Prompt for Taxonomy Classification (template)}
r"""You are a medical conversation scenario classifier.\\
\\
Input:\\
- question: the user’s full question (may include multiple turns).\\
\\
Task:\\
Classify the conversation along four independent axes and output labels with one-sentence definitions. Use plain language suitable for non-experts.\\
\\
Urgency: \{subcategory description\}\\
Uncertainty: \{subcategory description\}\\
Expertise: \{subcategory description\}\\
Task Type: \{subcategory description\}\\
- Assign exactly one primary\_task\_family capturing the main deliverable/intent, and optionally 0--2 secondary\_task\_families for important supporting needs.\\
\\
Output format (STRICT):\\
Return ONLY valid JSON with the following exact top-level schema (no extra keys, no markdown, no surrounding text):\\
\{\\
  "scenario": \{\\
    "urgency": \{"label": "...", "definition": "..."\},\\
    "uncertainty": \{"label": "...", "definition": "..."\},\\
    "expertise": \{"label": "...", "definition": "..."\},\\
    "primary\_task\_family": \{"label": "...", "definition": "..."\},\\
    "secondary\_task\_families": [\{"label": "...", "definition": "..."\}],\\
    "description": "..."\\
  \}\\
\}\\
\\
Now generate the output for:\\
question: \{\{question\}\}\\
""".strip()
\end{AIbox}
\caption{Prompt used to classify each question into taxonomy labels for principle retrieval and rubric generation.}
\label{fig:question_cls_prompt}
\end{figure*}

\paragraph{Notes.}
In the prompt, \{subcategory description\} is replaced with the label definitions for each axis from Appendix~\ref{sec:extract_task}, and the model is required to choose labels exactly as defined there.
We validate that returned labels are in-vocabulary and that the JSON schema is well-formed before extracting principles.

\section{Inference-time Rubric Guidance Tool}
\label{sec:principle_tool}

This appendix lists the prompts used by our inference-time tool. Given a question—optionally accompanied by prior dialogue context and/or a draft answer—the tool first classifies the scenario and then converts the retrieved principles into context-specific rubric items. In both prompts, {subcategory description} is instantiated with the taxonomy definitions in Appendix~\ref{sec:extract_task}. Importantly, all classifications and rubric generation are performed by the tool’s underlying model itself (i.e., without invoking any external models). The tool validates that predicted labels are in-vocabulary before extracting principles.

\begin{figure*}[t]
\begin{AIbox}{Prompt 1: Scenario Classification (with optional dialogue context and draft answer)}
r"""You are a medical conversation scenario classifier.\\
\\
Inputs (some may be empty):\\
- question: the user’s question\\
- dialogue\_context: prior messages (optional)\\
- draft\_answer: a candidate answer (optional)\\
\\
Task:\\
Use \{question\} as the primary signal. If dialogue\_context is provided, use it to clarify intent, constraints, and user expertise.\\
If draft\_answer is provided, use it only as weak evidence about what the assistant is attempting to do.\\
\\
Classify the scenario along four independent axes and output labels with one-sentence definitions. Use plain language suitable for non-experts.\\
\\
Urgency: \{subcategory description\}\\
Uncertainty: \{subcategory description\}\\
Expertise: \{subcategory description\}\\
Task Type: \{subcategory description\}\\
- Assign exactly one primary\_task\_family capturing the main deliverable/intent.\\
- Optionally assign 0--2 secondary\_task\_families for important supporting needs.\\
\\
Output format (STRICT):\\
Return ONLY valid JSON with the following exact schema (no extra keys, no markdown, no surrounding text):\\
\{\\
  "scenario": \{\\
    "urgency": \{"label": "...", "definition": "..."\},\\
    "uncertainty": \{"label": "...", "definition": "..."\},\\
    "expertise": \{"label": "...", "definition": "..."\},\\
    "primary\_task\_family": \{"label": "...", "definition": "..."\},\\
    "secondary\_task\_families": [\{"label": "...", "definition": "..."\}],\\
    "description": "..."\\
  \}\\
\}\\
\\
Now generate the output for:\\
question: \{\{question\}\}\\
dialogue\_context: \{ \{dialogue context\} \} \\
draft\_answer: \{\{draft answer\}\}\\
""".strip()
\end{AIbox}
\caption{Prompt used to classify a question when optional dialogue context and/or a draft answer is available.}
\label{fig:principle_tool_cls_prompt}
\end{figure*}

\begin{figure*}[t]

\begin{AIbox}{Prompt 2: Convert Retrieved Principles into Context-specific Rubric Items}
r"""You are generating scorable rubric items for evaluating a medical answer.\\
\\
Inputs:\\
- question: the user’s question\\
- dialogue\_context: prior messages (optional)\\
- draft\_answer: a candidate answer (optional)\\
- scenario: predicted taxonomy labels\\
- principles: retrieved HealthPrinciples matched to the scenario\\
\\
Task:\\
Convert the retrieved principles into rubric items that are tailored to this specific question and context.\\
If draft\_answer is provided, write rubric items that directly help diagnose and revise likely failures in the draft.\\
\\
Hard requirements:\\
- Output rubric items in ENGLISH.\\
- Each rubric item MUST be one sentence, specific, and checkable by a non-expert.\\
- Avoid vague style-only advice.\\
- Do NOT add new medical requirements beyond the provided principles and the user context.\\
- Prefer safety-critical items first when urgency is high or uncertainty is irreducible.\\
\\
Output format (STRICT):\\
Return ONLY valid JSON with the following exact schema (no extra keys, no markdown, no surrounding text):\\
\{\\
  "rubric\_items": ["...", "...", "..."],\\
  "principle\_coverage": [\\
    \{"principle\_id": "...", "used\_in\_items": [0, 2]\}\\
  ]\\
\}\\
\\
Now generate the output for:\\
question: \{\{question\}\}\\
dialogue\_context: \{\{dialogue\_context\}\}\\
draft\_answer: \{\{draft\_answer\}\}\\
scenario: \{\{scenario\_json\}\}\\
principles: \{\{principles\_json\}\}\\
""".strip()
\end{AIbox}
\caption{Prompt used to convert retrieved HealthPrinciples into question-specific rubric items at inference time.}
\label{fig:principle_tool_rubric_prompt}
\end{figure*}

\end{document}